\pdfoutput=1

\documentclass[11pt]{article}

\usepackage[]{EMNLP2022}

\usepackage{times}
\usepackage{latexsym}
\usepackage{array}
\usepackage{booktabs}
\usepackage{multirow}
\usepackage{amsmath}
\usepackage{amssymb}
\usepackage{graphicx}
\usepackage{enumerate}
\usepackage{diagbox}

\usepackage{color}

\newcommand{\Red}[1]{\textcolor[rgb]{1.00,0.00,0.00}{#1}}
\newcommand{\Blue}[1]{\textcolor[rgb]{0.00,0.00,1.00}{#1}}
\newcommand{\Green}[1]{\textcolor[rgb]{0.00,0.60,0.00}{#1}}

\newcommand{\Coral}[1]{\textcolor[rgb]{0.97,0.50,0.32}{#1}}

\newcommand{\finalcopy}[1]{{#1}}

\usepackage[T1]{fontenc}

\usepackage[utf8]{inputenc}

\usepackage{microtype}
\newcommand{\tabincell}[2]{\begin{tabular}{@{}#1@{}}#2\end{tabular}} 

\title{PseudoReasoner: Leveraging Pseudo Labels \\ for Commonsense Knowledge Base Population }

\author{Tianqing Fang$^{1}$, Quyet V. Do$^{1}$, Hongming Zhang$^2$, Yangqiu Song$^{1}$\\
{\bf Ginny Y. Wong$^3$,} {\bf Simon See$^3$,} \\
$^{1}$Department of Computer Science and Engineering, HKUST, Hong Kong SAR \\
$^{2}$Tencent AI Lab, Bellevue, USA\\
$^{3}$NVIDIA AI Technology Center (NVAITC), NVIDIA, Santa Clara, USA\\
 \texttt{\{tfangaa, vqdo, yqsong\}@cse.ust.hk} \\ \texttt{hongmzhang@global.tencent.com}, \texttt{\{gwong, ssee\}@nvidia.com}
\\
}

\begin{document}
\maketitle
\begin{abstract}

Commonsense Knowledge Base (CSKB) Population aims at reasoning over unseen entities and assertions on CSKBs, and is an important yet hard commonsense reasoning task. 
One challenge is that it requires out-of-domain generalization ability as the source CSKB for training is of a relatively smaller scale (1M) while the whole candidate space for population is way larger (200M). 
We propose PseudoReasoner, a semi-supervised learning framework for CSKB population that uses a teacher model pre-trained on CSKBs to provide pseudo labels on the unlabeled candidate dataset for a student model to learn from. 
The teacher can be a generative model rather than restricted to discriminative models as previous works.
In addition, we design a new filtering procedure for pseudo labels based on influence function and the student model's prediction to further improve the performance.
The framework can improve the backbone model KG-BERT (RoBERTa-large) by 3.3 points on the overall performance and especially, 5.3 points on the out-of-domain performance, and achieves the state-of-the-art. 
\finalcopy{Codes and data are available at \url{https://github.com/HKUST-KnowComp/PseudoReasoner}.}

\end{abstract}

\section{Introduction}

Commonsense knowledge are the common agreements by most people on daily entities, which are crucial for intelligent systems to act sensibly in the real world~\cite{DBLP:books/daglib/davis1990commonsense, liu2004conceptnet}. 
Endowing natural language understanding systems with the ability to draw commonsense reasoning remains an important yet challenging task. 

\begin{figure}[t]
    \centering
    \includegraphics[width=1.0\linewidth]{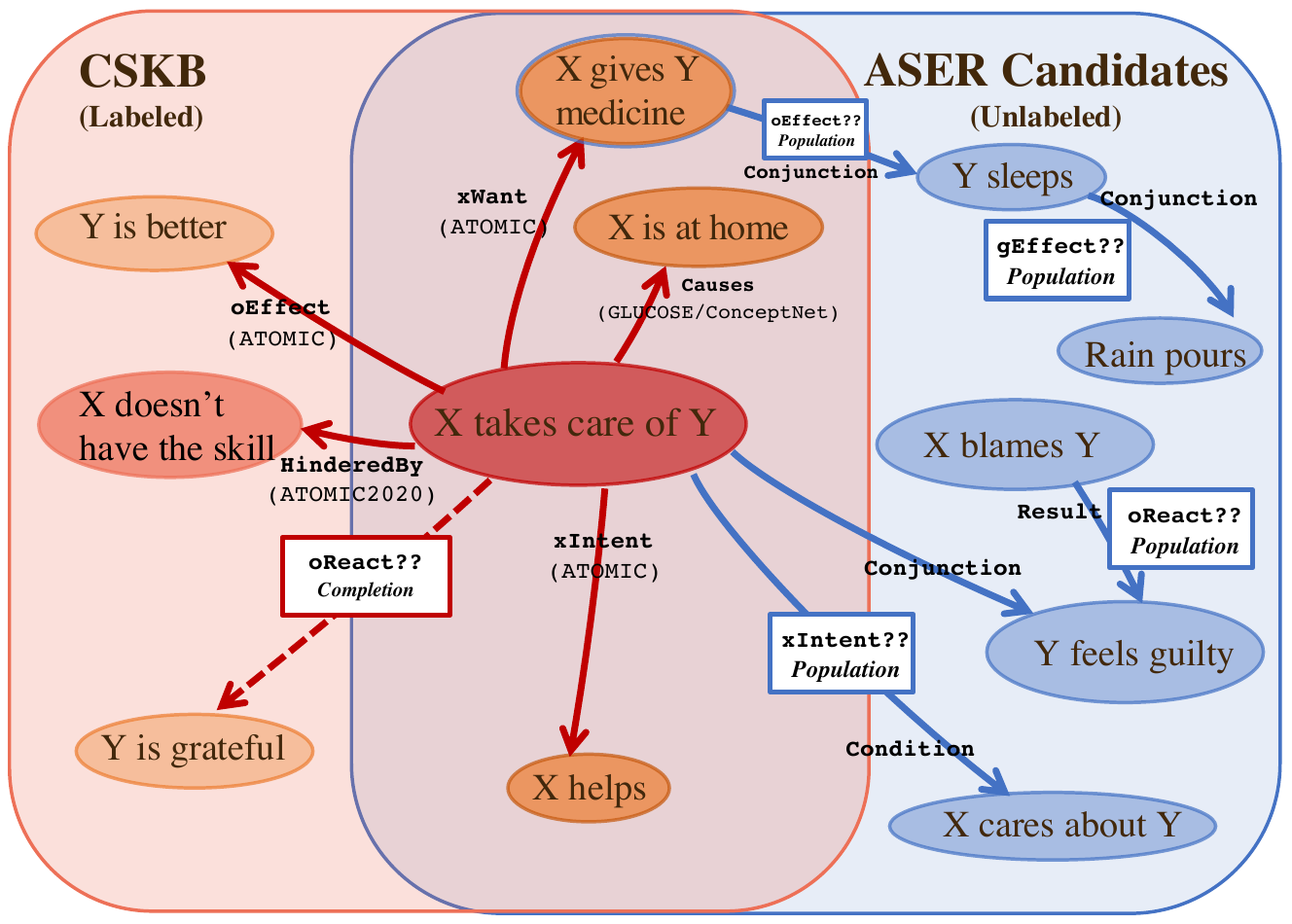}
    \vspace{-0.3in}
    \caption{An example of CSKB Population. The \Coral{\textbf{coral}} part (left) and the \Blue{\textbf{blue}} part (right) respectively represent the labeled CSKBs and the unlabeled candidate pool. The entities in the overlap parts are marked with coral shape and blue outline. The reasoning within the CSKB (coral-outlined boxes) belongs to the CSKB Completion part, and the reasoning that are not limited within the domain of CSKBs belongs to CSKB Population. }
    \label{fig:intro}
    \vspace{-0.1in}
\end{figure}

Throughout the development of automated commonsense understanding, CommonSense Knowledge Base (CSKB) is an important form of automatic commonsense reasoning system to store knowledge sources for drawing inferences. 
With expert-curated relations and human annotations, CSKBs such as ConceptNet~\cite{liu2004conceptnet}, ATOMIC~\cite{sap2019atomic, DBLP:conf/aaai/Hwang2021comet}, and GLUCOSE~\cite{mostafazadeh2020glucose} are developed to study commonsense regarding properties of objects, causes and effects of events and activities, motivations and emotional trajectories of humans on certain circumstances, and so on. 
As those human-annotated CSKBs are sparse and usually of a small scale and coverage, reasoning tasks on CSKB such as \textit{CSKB Completion}~\cite{DBLP:conf/acl/Li2016CKBC, DBLP:conf/conll/Saito2018CKBCG, DBLP:conf/aaai/Malaviya2020CKBC} and \textit{CSKB Population}~\cite{DBLP:conf/www/Fang2021discos, DBLP:conf/emnlp/Fang2021CKBP} are defined with the goal of either adding new edges/assertions within the training knowledge base (\textit{CSKB Completion}), or adding new edges/assertions from outside of CSKBs (\textit{CSKB Population}). A visualized comparison between the two tasks is shown in Figure~\ref{fig:intro}.

Different from CSKB Completion, which adopts a \textit{close-world assumption} and assume all knowledge are in-domain, the population task deals with unseen entities and requires a more out-of-distribution reasoning ability.
In this paper, we study commonsense reasoning in the context of CSKB Population. 
In this task, four mainstream CSKBs, ConceptNet, ATOMIC, ATOMIC$_{20}^{20}$, and GLUCOSE are aligned together as the labeled dataset. ASER~\cite{ZHANG2022103740aser}, a large-scale eventuality (events, activities, and states) knowledge graph is aligned with CSKBs and serves as unlabeled candidates for populating commonsense knowledge.
Human annotations on held-out dev/test sets sampled from both CSKBs and ASER are provided as the evaluation set.

There are two major challenges remaining unsolved for CSKB Population. First, the scale of the annotated training set (ConceptNet, ATOMIC, and GLUCOSE) is approximately 1M samples, too small compared with 200M of the actual candidate space to perform population (ASER). Second, as inherently only ground-truth (positive) examples are provided by CSKBs, the randomly sampled negative examples in the task are less informative and may lead the model to overfit artifacts of the dataset. 
A supervised learning model finetuned on such annotated training set is hard to be generalized to out-of-domain knowledge space, as shown in \citet{DBLP:conf/emnlp/Fang2021CKBP} and also Table~\ref{table:overall_result} in our paper, where the AUC for out-of-domain test sets performs over 10 points worse than the in-domain part. 

To address the above challenges, we propose \textbf{PseudoReasoner}, a semi-supervised learning framework that uses a pre-trained commonsense teacher model to automatically label the unlabeled candidates to serve as pseudo labels, such that the student model can be further finetuned with pseudo labels to improve out-of-domain commonsense reasoning ability. 
In fact, pre-trained language models finetuned on commonsense knowledge bases have shown to perform generalizable commonsense reasoning on downstream tasks to some extent. 
For example, leveraging commonsense knowledge generated by COMET~\cite{DBLP:conf/acl/Bosselut2019comet}, a language model finetuned on ATOMIC, can improve the performance on commonsense QA~\cite{DBLP:conf/emnlp/YangMFSBWBCD20GDAUG, DBLP:conf/emnlp/ShwartzWBBC20, DBLP:conf/aaai/BosselutBC21}.
Different from the text generation paradigm as in previous works, here we leverage the commonsense language model as a teacher model for labeling unlabeled candidates.
To further improve the quality of pseudo labels, we use both influence function~\cite{DBLP:conf/icml/KohL17} and the student model's prediction to select highly confident pseudo examples.

Our contribution is three-fold:


\vspace{-0.1in}

\begin{enumerate}[1)]
    \item We introduce a new way of providing pseudo label for CSKB Population by leveraging generative commonsense language models. \vspace{-0.1in}
    \item We propose a semi-supervised learning framework with pseudo labels and a special filtering mechanism based on influence function and student model's prediction that significantly improve the performance of CSKB Population, especially for out-of-domain knowledge triples. \vspace{-0.2in}
    \item We demonstrate the effectiveness of our framework by extensive experiments on different backbone models and different semi-supervised learning methods. We achieve the state-of-the-art performance on this task. 
\end{enumerate}

\vspace{-0.1in}
\section{Related Works}

\subsection{Commonsense Reasoning over Knowledge Bases}

Commonsense Knowledge Bases (CSKBs) provides rich human-curated commonsense knowledge in the form of head-relation-tail triples and are perfect fields to conduct commonsense reasoning. Reasoning tasks defined over CSKBs can be categorized into two main categories, CSKB Completion and CSKB Population. CSKB Completion adopts a close-world assumption and assume the knowledge not included in the knowledge base are incorrect, thus using evaluation metrics such as accuracy, MRR, and HITS@k (under the link prediction setting~\cite{DBLP:conf/acl/Li2016CKBC, DBLP:conf/aaai/Malaviya2020CKBC, DBLP:conf/conll/Saito2018CKBCG, jastrzebski-etal-2018-commonsense}) or BLEU (under the text generation setting~\cite{DBLP:conf/acl/Bosselut2019comet, DBLP:conf/aaai/Hwang2021comet}) to evaluate the reasoning performance. 
As vanilla knowledge base completion models such as TransE~\cite{NIPS2013_transe} and ComplEx~\cite{DBLP:conf/icml/TrouillonWRGB16complex} do not take the node semantics into account and CSKBs are much sparser than standard factual KBs, \citet{DBLP:conf/aaai/Malaviya2020CKBC} and \citet{DBLP:conf/ijcnn/Wang2021inductive} use BERT~\cite{DBLP:conf/naacl/DevlinCLT19bert} to encode nodes and propose graph densifiers to address the sparsity issue.

CSKB Population, on the other hand, requires the model to reason on not only existing nodes in the CSKB~\cite{DBLP:conf/www/Fang2021discos, DBLP:conf/emnlp/Fang2021CKBP}, but the nodes from other domains (e.g., ASER~\cite{ZHANG2022103740aser}). The evaluation set is manually annotated to form a binary classification task, and AUC is used as the evaluation metric. 
The population task is inherently a classification task, KG-BERT~\cite{DBLP:journals/corr/yao2019kgbert} and KG-BERTSage~\cite{DBLP:conf/emnlp/Fang2021CKBP}, a graph-aware model based on KG-BERT, are used to tackle the task.
\finalcopy{While there are other ways of populating commonsense knowledge, such as leveraging conceptualization~\cite{he2022acquiring} or generative language models~\cite{DBLP:conf/aaai/BosselutBC21, DBLP:conf/naacl/WestBHHJBLWC22}, they are different with CSKB Population in several aspects. 
For conceptualization, they provide new triples in depth through hierarchical structures of concepts by lexical manipulation and the core semantics are preserved, while we do population in breadth by exploring more diverse information-extracted triples. 
For text generation, first, text generation itself doesn't perform reasoning, while discrimination on a knowledge graph can provide structural and contextual information to do reasoning. 
Second, generative models are trained only on positive/plausible examples by nature while do not handle negative/implausible examples. 
We will show the importance of negative examples in CSKB Population in the experiments. }

\begin{table}[t]
\small
\centering
\begin{tabular}{c|cc}
\toprule
& $D_l$ & $D_u$ \\
\midrule
\# triples & 1,119,517 (training) & 218,809,746 \\
\bottomrule
\end{tabular}
\vspace{-0.1in}
\caption{ Statistics of labeled $D_l$ (CSKB with negative examples) and unlabeled $D_u$ (processed ASER). }
\vspace{-0.2in}
\label{table:stat}
\end{table}

\vspace{-0.1in}
\subsection{Pseudo Labels}

Pseudo labels are widely used in the semi-supervised learning setting~\cite{ DBLP:conf/cvpr/IscenTAC19,DBLP:conf/cvpr/XieLHL20,DBLP:conf/nips/SohnBCZZRCKL20,DBLP:conf/cvpr/PhamDXL21}.
In common, pseudo labels alleviates the cost of human annotation, or familiarizes the student model with out-of-domain or same-distribution-but-unseen data (i.e data from the large unlabeled dataset) under the advice from teacher models.
It has been used in image classification~\cite{lee2013pseudo, DBLP:conf/eccv/ShiGDMTZ18, DBLP:journals/corr/abs-1905-00546, DBLP:conf/cvpr/XieLHL20, DBLP:conf/cvpr/PhamDXL21}, machine translation~\cite{DBLP:conf/iclr/HeGSR20, DBLP:conf/emnlp/ChenZKM021}, information retrieval~\cite{DBLP:journals/corr/abs-2112-07577}, and word segmentation~\cite{DBLP:conf/acl/HuangLHXLS21}. 
In the simplest form of pseudo labels, the teacher model that provides pseudo labels is static or iteratively updated using the latest student model. Meta pseudo label~\cite{DBLP:conf/cvpr/PhamDXL21}, on the other hand, leverages meta-learning technique to learn an end-to-end teacher-student network that updates both networks jointly based on the performance of the student network on the labeled dataset, yielding state-of-the-art performance on ImageNet~\cite{DBLP:conf/cvpr/DengDSLL009}. 

\subsection{Other Semi-supervised Learning Methods}


Most works on semi-supervised learning leverages the idea of consistency training~\cite{DBLP:conf/nips/ZhouBLWS03, DBLP:conf/nips/XieDHL020UDA, DBLP:conf/acl/GururanganDCS19}, which aims to constrain the model to be robust given noised inputs or hidden states. UDA (Unsupervised Data Augmentation)~\cite{DBLP:conf/nips/XieDHL020UDA} adopts back translation and TF-IDF word replacement to provide noisy inputs and use them for consistency training. 
However, in the setting of CSKB Population, the problem of out-of-distribution is more important as we aim to explore novel commonsense knowledge that are not seen in the training domain, which makes directly finetuning on pseudo labels more effective than using consistency training.


\section{CSKB Population}

\subsection{Problem Definition}

Denote a labeled ground-truth Commonsense Knowledge Base as $D^{+}_l = \{(h, r, t)| h\in H_C, r \in R, t \in T_C \}$. The overall labeled dataset $D_l=\{(h, r, t), y)\}$ is composed of $D^{+}_l$ and a randomly sampled negative dataset $D^{-}_l$ from the CSKB, where $y\in\{0, 1\}$ is the label of the triple. Triples from $D^{+}_l$ are labeled 1 while those from $D^{-}_l$ are labeled 0.
$H_C$ and $T_C$ are the set of heads and tails in the CSKB. $R$ is the relation set.

$D_u = \{(h, r, t)| h\in H_u, r \in R, t \in T_u\}$ is an unlabeled candidate knowledge base of the same format and relation set as the CSKB. It is of a much larger scale than the labeled part and is a source for populating commonsense knowledge.
$H_u$ and $T_u$ is the set of heads and tails in the unlabeled KB.
The task is defined as given the labeled commonsense knowledge base $D_l$ as training source, predict the plausibility of triples from  $D_u$.


\begin{table}[t]
\scriptsize
\centering
\begin{tabular}{c|ccc}
\toprule
&\tabincell{c}{\scriptsize{\textit{Original}}\\ \scriptsize{\textit{Test Set}}} & \tabincell{c}{\scriptsize{\textit{CSKB head} }\\ \scriptsize{\textit{+ ASER tail}}} & \tabincell{c}{\scriptsize{\textit{ASER}}\\ \scriptsize{\textit{edges}}}\\
\midrule
\tabincell{c}{Sampled from} & $D_l$ & $D_l$ head, $D_u$ tail & $D_u$  \\ 
Domain & In-domain & Out-of-domain & Out-of-domain  \\ 
\# triples (dev) & 2,042 & 2,193 & 1,982\\
\# triples (test) & 8,437 & 9,103 & 7,974 \\
\bottomrule
\end{tabular}
\vspace{-0.1in}
\caption{ The details of evaluation set categorization. }
\vspace{-0.2in}
\label{table:intro_eval_set}
\end{table}

We use the dataset provided by \citet{DBLP:conf/emnlp/Fang2021CKBP}.
Here, the set of heads $H_C$, tails $T_C$, and relations $R$ come from the alignment of ConceptNet, ATOMIC, ATOMIC$_{20}^{20}$ (newly developed relations), and GLUCOSE, as in the original paper. 
The unlabeled KB $D_u$ is adapted from ASER, where the discourse relations are converted to commonsense relations to serve as candidates for population. 
The evaluation dataset with 32K triples is sampled from both $D_l$ and $D_u$ and manually annotated. 
There are three categories of the evaluation set, \textit{Original Test Set}
, \textit{CSKB head + ASER tail}, and \textit{ASER edges}, where the first category is sampled from the held-out test split in $D_l$ (both $D_l^+$ and negative examples $D_l^-$) and is thus an in-domain test set, and the latter two are novel assertions outside of $D_l$ and are thus out-of-domain. The statistics and descriptions of the training and evaluation datasets are shown in Table~\ref{table:stat} and~\ref{table:intro_eval_set}. The list of commonsense relations, alignment methods, and more detailed statistics of the dataset can be found in the Appendix~\ref{appendix:ckbp_task}.

\begin{figure*}[t]
    \centering
    \includegraphics[width=0.9\linewidth]{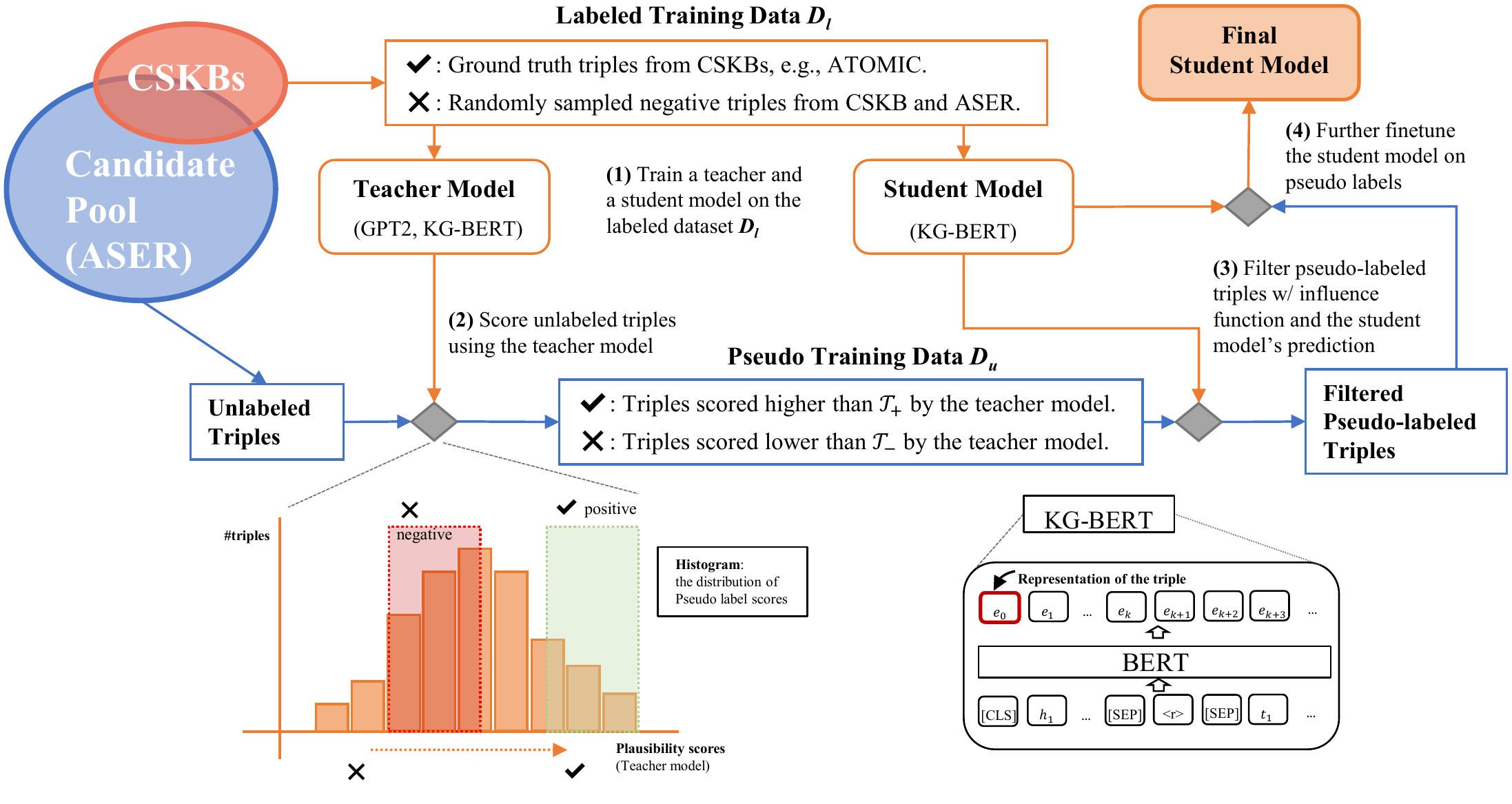}
    \vspace{-0.1in}
    \caption{An end-to-end workflow of \textbf{PseudoReasoner}. Four steps in the figure are elaborated in Section~\ref{sec:methods}.}
    \label{fig:model_sketch}
    \vspace{-0.15in}
\end{figure*}

\subsection{Backbone Models}

Considering the nature of the CSKB Population task is triple classification in the form of natural language, we use KG-BERT~\cite{DBLP:journals/corr/yao2019kgbert} as the backbone model following~\citet{DBLP:conf/emnlp/Fang2021CKBP}. 
In detail, a triple $(h, r, t)$ is concatenated and serialized as ``$\text{[CLS]}, h_1, ..., h_{|h|}, \text{[SEP]}, \text{[$r$]}, \text{[SEP]}, t_1, ..., t_{|t|}$''. Here, [CLS] and [SEP] are the special tokens in BERT-based models~\cite{DBLP:conf/naacl/DevlinCLT19bert}. [CLS] is used to represent the whole sentence, and [SEP] is used to separate different sentences, respectively. $h_1, ..., h_{|h|}$ are the tokens of the head $h$, and  $t_1, ..., t_{|t|}$ are the tokenized tokens of the tail $t$. [$r$] is registered as a new special token for a certain relation $r$. After feeding the serialized version of $(h, r, t)$ into a BERT-based masked language model, the representation of the special token [CLS] is regarded as the representation of the whole triple. 
It is trained to distinguish positive triples with negative triples with cross entropy loss. Here $x$ denotes a triple $(h, r, t)$, $P_L$ models the distribution of the labeled dataset $D_l$, 
and $\theta$ is the set of parameters for KG-BERT. $P_{\theta}(y|x)$ denotes the probability after feeding the model prediction logits to softmax under parameter set $\theta$. Then the optimization objective is as follows:
\vspace{-0.2in}

\begin{equation}\label{eq:kgbert_objective}
    J(\theta) = \mathbb{E}_{x_l \sim P_L(x)} [ - \log P_\theta(y|x_l)]. \\
\end{equation}

\vspace{-0.1in}

\section{Methods}\label{sec:methods}


In this section, we present the details of the framework of PseudoReasoner. A sketch illustration of the model is presented in Figure~\ref{fig:model_sketch}. To sum up, the procedure of PseudoReasoner can be summarized into the following steps:

\begin{enumerate}[1)]
    \item Train a teacher model and a student model on the labeled dataset $D_l$ (Section~\ref{sec:teacher}). \vspace{-0.1in}
    \item Use the teacher model to predict plausibility scores on triples from the unlabeled $D_u$. Triples with high/low plausibility scores within pre-defined intervals are given label $1/0$ (Section~\ref{sec:pseudo_label}). \vspace{-0.1in}
    \item Filter the pseudo labels with influence function with respect to the student model, and the student model's predictions.  (Section~\ref{sec:filter}). \vspace{-0.1in}
    \item Finetune the student model on filtered pseudo labels from 3). \vspace{-0.1in} (Section~\ref{sec:pseudoreasoner_training}).
\end{enumerate}
\vspace{-0.1in}

\subsection{Pseudo Label Construction}

\subsubsection{Teacher Models}\label{sec:teacher}

We use a teacher model pre-trained on the labeled dataset for labeling the unlabeled triples. We define plausibility scores of an unlabeled triple $x$ as $\alpha(x)$, where the higher the score the more plausible the triple is regarded by the teacher model. We choose two different forms of teacher models as follows:

\vspace{0.5em}

\noindent \textbf{GPT2}~\cite{radford2019gpt2}: As negative sampling in the labeled dataset $D_l$ is noisy, we aim to use an alternative model that avoids the negative part $D_l^-$.
We finetune a (COMET) GPT2 language model, as the representative of generative family, on the positive part of the labeled dataset, $D_l^+$, with a text generation task. 
For an $(h, r, t)$ triple from $D_l^+$, denote $x$ as the serialized version of the triple, ``$h_1, ..., h_{|h|}, \text{[$r$]}, t_1, ..., t_{|t|}$''. 
$\theta_{LM}$ denotes the trainable parameters in GPT2 language model (LM).
We minimize the negative log likelihood of each triple as indicated in Equation~(\ref{eq:gpt2loss}):

\vspace{-0.2in}
\begin{equation}\label{eq:gpt2loss}
    L(x, \theta_{LM}) = -\frac{1}{|x|}\sum_{i=1}^{|x|} \log P(x_i|x_{<i}, \theta_{LM}).\\
\end{equation}

Denote the optimized parameters as $\theta^*_{LM}$. Here, the plausibility function $\alpha(x) = -L(x, \theta_{LM}^*)$, where the lower the loss, the higher the plausibility score by GPT2.
Hence, for the triples from the unlabeled dataset, $\mathcal{D}_u$, we score every triple with Equation~(\ref{eq:gpt2loss}) on $\theta_{LM}^*$.


\vspace{0.5em}

\noindent\textbf{KG-BERT}: Besides GPT2, KG-BERT itself, a discriminative model, can be used as a teacher model. This teacher model learns $\theta^*$ from the labeled dataset $D_l$ with cross entropy loss in Equation~(\ref{eq:kgbert_objective}).
For an instance $\{(h, r, t), y\}\in D_l$, denote $x = (h, r, t)$, we use $\alpha(x)$ $= P_{\theta^*}(y\text{=}1|x)$ as $x$'s plausibility score. 


\begin{table*}[t]
\small
\centering
\renewcommand\arraystretch{1.1}
\begin{tabular}{l|l|c|c|cc|cc}
\toprule
\multirow{3}{*}{Category} & \multirow{3}{*}{Model} & \multirow{3}{*}{all} & In-domain & \multicolumn{2}{c|}{Out-of-domain (OOD)}  & \multirow{3}{*}{$\Delta_{all}$} & \multirow{3}{*}{\scriptsize{$\Delta_{OOD}$}}\\ \cline{4-6}
& & & \tabincell{c}{\textit{Original}\\ \textit{Test Set}} &\tabincell{c}{\textit{CSKB head }\\ \textit{+ ASER tail}} & \tabincell{c}{\textit{ASER}\\ \textit{edges}} & & \\
\midrule

\multirow{12}{*}{\tabincell{l}{Supervised\\ Learning}} & KG-BERT (BERT-base) \scriptsize{\textit{110M}} & 62.5 & 74.2 & 51.9 & 54.7 & - & - \\
& KG-BERT (BERT-large) \scriptsize{\textit{340M}} & 67.7 & 74.5 & 58.7 & 62.0 & - & -  \\
& KG-BERT (DeBERTa-base) \scriptsize{\textit{100M}} & 64.5 & 73.2 & 54.0 & 57.0 & - & -  \\
& KG-BERT (DeBERTa-large) \scriptsize{\textit{350M}} & 69.2 & 77.6 & 59.9 & 61.8 & - & -  \\
& KG-BERT (BART-base) \scriptsize{\textit{139M}} & 65.1 & 74.7 & 54.7 & 56.6 & - & -  \\
& KG-BERT (BART-large) \scriptsize{\textit{406M}} & 70.4 & \underline{78.6} & 62.8 & 64.2 & - & -  \\
& KG-BERT (RoBERTa-base) \scriptsize{\textit{110M}} & 68.0 & 76.3 & 59.3 & 59.8 & - & - \\
& KG-BERT (RoBERTa-large) \scriptsize{\textit{340M}} & \underline{70.9} & 78.0 & \underline{63.4} & \underline{64.6} & - & - \\ \cline{2-6}
& COMET (GPT2-small) \scriptsize{\textit{117M}} & 69.6 & 71.6 & 67.4 & 65.0 & - & - \\
& COMET (GPT2-medium) \scriptsize{\textit{345M}} & 69.7 & 71.9 & 67.0 & 67.9 & - & - \\
& COMET (GPT2-large) \scriptsize{\textit{774M}} & 70.6 & 73.7 & 66.8 & 68.0 & - & - \\
& COMET (GPT2-XL) \scriptsize{\textit{1558M}} & 70.7 & 74.6 & 66.7 & 67.6 & - & - \\
\midrule
\multirow{4}{*}{\tabincell{l}{Semi-supervised\\ Learning \\ (RoBERTa-large) }} 
& UDA (TF-IDF)  & 71.7 & 78.0 & 65.1 & 65.9 & +0.8 & +1.5\\
& UDA (back-trans.)  & 71.6 & 78.6 & 64.2 & 66.2 & +0.7 & +1.2\\
& G-DAUG  & 71.7 & 78.5 & 64.8 & 65.5 & +0.8 & +1.2 \\
& \finalcopy{G-DAUG (COMET-distill)}  & \finalcopy{72.2} & \finalcopy{78.6}& \finalcopy{65.9}&\finalcopy{66.9}&+1.3&+2.4  \\
& Noisy-student  & 72.4 & 79.3 & 65.3 & 66.7 & +1.5 & +2.0\\
\midrule
\multirow{2}{*}{Ours} & \textbf{PseudoReasoner} (BERT-base) & 67.9 & 76.0 & 56.1 & 64.2 & +5.4 &  +6.9 \\
& \textbf{PseudoReasoner} (RoBERTa-large) & \textbf{74.2} & \textbf{80.1} & \textbf{69.5} & \textbf{69.3} & +3.3 & +5.3 \\
\bottomrule
\end{tabular}
\caption{ Results on the test set of the CSKB Population benchmark. For supervised learning baselines, we report the result of KG-BERT~\cite{DBLP:journals/corr/yao2019kgbert} with four backbone encoders and GPT2 (use LM loss to score triples). For semi-supervised learning (SSL) baselines, we study UDA~\cite{DBLP:conf/nips/XieDHL020UDA}, G-DAUG~\cite{DBLP:conf/emnlp/YangMFSBWBCD20GDAUG}, and Noisy-student~\cite{DBLP:conf/cvpr/XieLHL20}. The backbone encoders for SSL baselines are RoBERTa-large, which performs the best in the supervised setting. 
The number of parameters of backbone language models are presented as subscripts behind model names. $\Delta_{all}$ and $\Delta_{OOD}$ are the improvement on the ``all'' AUC and the Out-of-domain (OOD) AUC.}
\label{table:overall_result}
\vspace{-0.1in}
\end{table*}

\subsubsection{Acquiring Pseudo Labels}\label{sec:pseudo_label}

The triples whose plausibility scores $\alpha(x)$ are between $[\mathcal{T}^-_{min}, \mathcal{T}^-_{max}]$ are labeled as negative, and the triples within $[\mathcal{T}^+_{min}, \mathcal{T}^+_{max}]$ are labeled as positive. Here $\mathcal{T}^-_{min} < \mathcal{T}^-_{max} < \mathcal{T}^+_{min} < \mathcal{T}^+_{max}$. The reason that we introduce additional  $\mathcal{T}^-_{min}$ and $\mathcal{T}^+_{max}$ is that we want to filter out the triples that are treated over plausible or implausible by GPT2 to reduce potential \textit{selection bias}. For example, GPT2 has been shown to provide low loss for repetitive patterns instead of the plausibility of the semantics ~\cite{DBLP:conf/nips/BrownMRSKDNSSAA20}. Details of the hyperparameter selection are in Appendix~\ref{appendix:hyper_params}.

\subsubsection{Pseudo Label Filters}\label{sec:filter}

To further improve the quality of pseudo labels, we propose two filtering mechanisms on pseudo labels for better finetuning.

\noindent \textbf{Influence Function. } Filtering out detrimental training examples with influence function~\cite{DBLP:conf/icml/KohL17} can boost the model performance, as shown in~\citet{DBLP:conf/emnlp/YangMFSBWBCD20GDAUG} and \citet{ DBLP:conf/acl/HanWT20}.
A training example $z=((h, r, t), y)$ will hurt the generalization ability of the model if including $z$ in the training set results in a higher validation loss. Denote $L(\mathcal{Z}, \theta)$ as the loss function of dataset $\mathcal{Z}$ under the parameter set $\theta$. Then the loss under training set $\mathcal{Z}_{train}$ is indicated in Equation~(\ref{eq:influence_loss}): 
\vspace{-0.1in}
\begin{equation}\label{eq:influence_loss}
    L(\mathcal{Z}_{train}, \theta) = \frac{1}{|\mathcal{Z}_{train}|} \sum_{i=1}^{|\mathcal{Z}_{train}|} L(z_i, \theta).
\end{equation}
Denote $\theta^*$ as the optimized parameters after training the model on $\mathcal{Z}_{train}$, and $\theta^*_{-z}$ as the optimized parameters after training the model on $\mathcal{Z}_{train}-\{z\}$.
Denote $\mathcal{Z}_{val}$ as the validation set.
The empirical criterion to determine $z$ as a detrimental training example is Equation~(\ref{eq:influence_criterion}):
\vspace{-0.0in}
\begin{equation}\label{eq:influence_criterion}
    L(\mathcal{Z}_{val}, \theta^*) - L(\mathcal{Z}_{val}, \theta^*_{-z})>0.
\end{equation}
The left-hand-side $L(\mathcal{Z}_{val}, \theta^*) - L(\mathcal{Z}_{val}, \theta^*_{-z})$ can be approximated without retraining the model by influence function~\cite{DBLP:conf/icml/KohL17}:
\vspace{-0.2in}

\begin{equation}
\mathcal{I}_{up, loss}(z) = - \nabla_{\theta} L(z_{val}, \theta^*)^\top H_{\theta^*}^{-1} \nabla_{\theta} L(z, \theta^*) ,
\end{equation}
where $H_{\theta^*}=\frac{1}{|\mathcal{Z}_{train}|} \sum_{z_i \in \mathcal{Z}_{train}} \nabla^2_{\theta^*} L(z_i, \theta^*) $ is the Hessian. We linearly approximate $\mathcal{I}_{up, loss}$ with inverse hessian-vector product (HVP) introduced in LiSSA~\cite{DBLP:journals/jmlr/AgarwalBH17} following~\cite{DBLP:conf/icml/KohL17}.
Details about influence function and the numerical approximation to calculate it can be found in the Appendix~\ref{appendix:influence_function}.
We filter out those examples with negative influence scores, which are harmful to the generalization of the model.

\vspace{0.5em}

\noindent \textbf{KG-BERT.} As the student model we use is KG-BERT, when the pseudo labels from GPT2 are used, we can use the $P_{\theta^*}(y|x)$ produced from optimized KG-BERT as an additional filter to select pseudo labels. Specifically, pseudo labels $\{x=(h, r, t), y\}$ with $P_{\theta^*}(y|x)>0.5$ are selected. This procedure can be viewed as ensembling GPT2 and KG-BERT. 

\subsection{PseudoReasoner Training}\label{sec:pseudoreasoner_training}

The objective function of KG-BERT on the labeled dataset is shown in Equation~(\ref{eq:kgbert_objective}),
and the objective function on pseudo labels can be written as:
\vspace{-0.1in}
\begin{equation}\label{eq:pseudo_loss}
    J_U(\theta) = \mathbb{E}_{x_u \sim P_U(x)}\mathbb{E}_{\hat{y} \sim q(y|x_u)}  [ - \log P_\theta(\hat{y}|x_u)]. \\ 
\end{equation}
Here $P_L$ and $P_U$ are the distribution of the labeled and unlabeled dataset, respectively. $q(y|x)$ is the distribution of pseudo labels, modeled by the teacher model and filters. After finetuning KG-BERT initialized with $\theta^*$ on the filtered pseudo labels with Equation~(\ref{eq:pseudo_loss}), we acquire ${\theta^*}'.$

\vspace{-1em}

\section{Experiments}
\vspace{-0.5em}
\subsection{Baselines}

For the supervised learning setting, we use KG-BERT~\cite{DBLP:journals/corr/yao2019kgbert} and COMET~\cite{DBLP:conf/acl/Bosselut2019comet} (GPT2) to perform CSKB Population. 
For KG-BERT, as it's flexible to be adapted using different pre-trained encoders, we use BERT~\cite{DBLP:conf/naacl/DevlinCLT19bert}, RoBERTa~\cite{DBLP:journals/corr/abs-1907-11692roberta}, DeBERTa~\cite{DBLP:conf/iclr/HeLGC21deberta}, and BART~\cite{DBLP:conf/acl/LewisLGGMLSZ20bart} as the backbone language models. For BERT, RoBERTa, and DeBERTa, we use the embedding of the [CLS] token in KG-BERT as the representation of the whole triple. For BART, we follow the ways of doing sequence classification in the original paper~\cite{DBLP:conf/acl/LewisLGGMLSZ20bart} to use the embedding of the end-of-sentence token in the decoder as the representation of the whole triple.

For the semi-supervised learning setting, we use the following baseline models:

\noindent \textbf{Unsupervised Data Augmentation (UDA). } UDA~
\cite{DBLP:conf/nips/XieDHL020UDA} uses consistency training to constrain the model to provide invariant predictions with noise added to the input. We adapt UDA into the framework of CSKB Population and uses TF-IDF word replacement and back-translation to provide noise to the input text to be fed into the consistency loss. More details are in the Appendix~\ref{appendix:uda}.

\noindent \textbf{Noisy Student.} Noisy student~\cite{DBLP:conf/cvpr/XieLHL20} 
trains a student model with noise added during training iteratively. 
A teacher model is first trained to provide hard or soft pseudo labels for a student model to finetune together with the labeled dataset. 
Soft pseudo labels mean using logit scores after softmax as labels. 
Then the student model is iteratively re-used as the teacher model and a new student model is acquired through each iteration.
Details can be found in the Appendix~\ref{appendix:noisy_student}.

\noindent \textbf{Generative Data Augmentation (G-DAUG).} 
G-DAUG~\cite{DBLP:conf/emnlp/YangMFSBWBCD20GDAUG} leverages text generation language models to automatically generate pseudo training data examples for finetuning. Though it's not designed for semi-supervised learning, we adapt it to our framework of pseudo labeling to serve as a semi-supervised learning baseline. 
We use COMET (GPT2-XL) finetuned on $D_l$ to generate pseudo examples with heads from $D_u$. Then those pseudo labels are filtered with both influence function, diversity heuristics, and KG-BERT scores. 
Then those pseudo examples are used the same way as in our PseudoReasoner for further finetuning. 
\finalcopy{We also try to replace COMET with COMET-distill~\cite{DBLP:conf/naacl/WestBHHJBLWC22}, the COMET trained with distilled commonsense knowledge from GPT3, which has a better performance and capacity than the vanilla COMET. }
Details can be found in the Appendix~\ref{appendix:gdaug}. 

\subsection{Experimental Settings}\label{sec:experimental_settings}

The learning rate for all models are set as 1e-5, and the batch size is 64. We use the framework of Huggingface Transformers\footnote{https://huggingface.co/} to form our codebase.
Early stopping is used where the best checkpoint is selected when the largest validation AUC is achieved. For all experiments, we report the average scores across three different random seeds. 

\finalcopy{For thresholding, we set the thresholds $\mathcal{T}^-_{min}\text{=}-4.0, \mathcal{T}^-_{max}\text{=}-3.7, \mathcal{T}^+_{min}\text{=}-2.8, \mathcal{T}^+_{max}\text{=}-2.0$, by roughly observing the data distribution and representative knowledge triples in different range of plausibility scores. We then randomly down-sample the pseudo examples such that the number is the same as the original training data. Ablations are provided in section~\ref{sec:threshold_ablation}}

See more details about thresholding and hyperparameters of different models in Appendix~\ref{appendix:hyper_params}.

\begin{table}[t]
\small
\centering
\begin{tabular}{l|c|ccc}
\toprule
  \tabincell{c}{Teacher models} & \tabincell{c}{all} & \tabincell{c}{\scriptsize{\textit{Original}}\\ \scriptsize{\textit{Test Set}}} & \tabincell{c}{\scriptsize{\textit{CSKB head} }\\ \scriptsize{\textit{+ ASER tail}}} & \tabincell{c}{\scriptsize{\textit{ASER}}\\ \scriptsize{\textit{edges}}} \\

\midrule
\multicolumn{1}{l|}{N/A (Baseline)}  & 70.9 & 78.0 & 63.4 & 64.6\\

\midrule
\multicolumn{2}{l}{\textbf{w/o all filters}} \\
\hline
RoBERTa-large  & 71.8 & 79.1 & 65.1 & 64.6 \\
GPT2-small  & 72.3 & 79.5 & 65.1 & 65.9 \\
GPT2-medium  & 72.6 & 79.5 & 65.8 & 68.0 \\
GPT2-XL  & 72.8 & 80.1 & 66.3 & 66.0 \\
\midrule
\multicolumn{2}{l}{\textbf{w/ all filters}} \\
\hline
RoBERTa-large  & 72.3 & 79.2 & 65.4 & 66.3 \\
GPT2-small  & 73.7 & 79.8 & 67.5 & 68.8 \\
GPT2-medium  & 74.1 & \textbf{81.3} & 68.1 & 69.0 \\
GPT2-XL  & \textbf{74.2} & 80.1 &  \textbf{69.5} & \textbf{69.3} \\

\bottomrule
\end{tabular}
\caption{Performance of \textbf{PseudoReasoner} (RoBERTa-large) using different teacher models w/ or w/o filters.} 
\label{table:ablation_teacher_model}
\vspace{-0.2in}
\end{table}

\subsection{Results}

The main results are shown in Table~\ref{table:overall_result}. We compare the results of both supervised learning and semi-supervised learning approaches with our proposed model PseudoReasoner. The ``all'' column presents the overall AUC across all testing examples and is the main metric of CSKB Population. We also separately present the AUC of different test set categories, In-domain (\textit{Original Test Set}) and Out-of-domain (\textit{CSKB head + ASER tail}, and \textit{ASER edges}). In the last two columns, we report the increase by applying different semi-supervised learning approaches under the same backbone model, where $\Delta_{all}$ means the increase of ``all'' (AUC) metric and $\Delta_{OOD}$ means the increase of AUC for out-of-domain test sets.

For supervised-learning approaches, KG-BERT-based models mostly perform well on the In-domain test set while have poorer generalization ability to Out-of-domain test sets compared to COMET (GPT2). As GPT2 is only finetuned on the positive part of the dataset, it suffers less from the bias of negative sampling in $D_l^-$ and has a better generalization on new knowledge. However, the drawback in In-domain reasoning hinders the overall performance of GPT2-XL from surpassing KG-BERT (RoBERTa-large), even with 4.5 times more parameters.

For semi-supervised learning baselines, all of them can increase the performance of the backbone KG-BERT (RoBERTa-large), especially for the out-of-domain split. However, the improvement on the In-domain part remains insignificant and the improvement on out-of-domain part is not as competitive as our PseudoReasoner. As we use the same code base and training method to train all semi-supervised learning methods, the main differences between PseudoReasoner and other SSL methods lie in the ways of processing the unlabeled dataset. We leave the detailed discussions in the next section (Section~\ref{section:comparison}).

\begin{table}[t]
\small
\centering
\begin{tabular}{l|c|ccc}
\toprule
 Filter & \tabincell{c}{all} & \tabincell{c}{\scriptsize{\textit{Original}}\\ \scriptsize{\textit{Test Set}}} & \tabincell{c}{\scriptsize{\textit{CSKB head} }\\ \scriptsize{\textit{+ ASER tail}}} & \tabincell{c}{\scriptsize{\textit{ASER}}\\ \scriptsize{\textit{edges}}} \\
\midrule
baseline & 70.9 & 78.0 & 63.4 & 64.6 \\
\ w/o filter & 72.8 & 80.1 & 66.3 & 66.0\\
\ + influence & 73.2 & 78.0 & 68.3 & \textbf{70.6} \\
\ + KG-BERT & 73.7 & 79.5 & 68.9 & 68.6 \\
\ + both & \textbf{74.2} & \textbf{80.1} & \textbf{69.5} & 69.3 \\
\bottomrule
\end{tabular}
\caption{ The effects of different filters on pseudo labels when KG-BERT (RoBERTa-large) is the backbone and GPT2-XL is the teacher model.
}
\label{table:ablation_filter}
\vspace{-0.3in}
\end{table}

\section{Analysis and Discussions}

In this section, we discuss the ablation study on model components, the comparisons with semi-supervised learning baselines, diversity analysis, and discussions about why PseudoReasoner works.

\begin{figure}[t]
    \centering
    \includegraphics[width=0.7\linewidth]{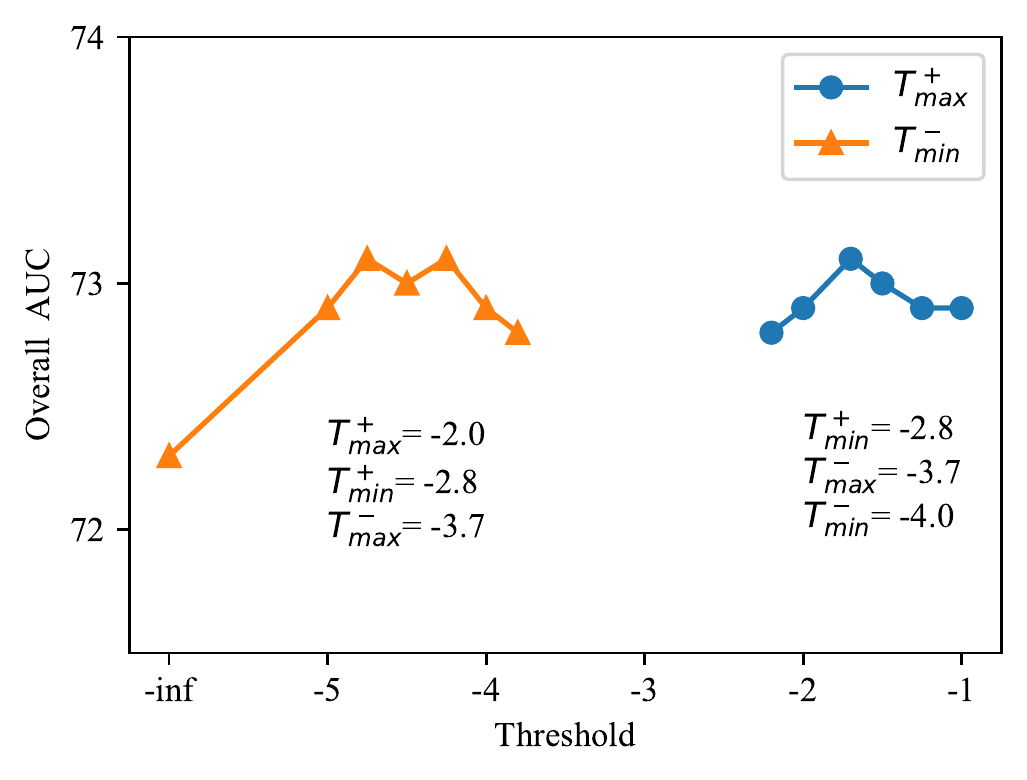}
    \vspace{-0.16in}
    \caption{Ablation study on different $\mathcal{T}_{max}^+$ and $\mathcal{T}_{min}^-$.}
    \label{fig:threshold_ablation}
    \vspace{-0.2in}
\end{figure}

\subsection{Ablation Study}
We study the effects of different teacher models and the filters on pseudo labels.

\subsubsection{Teacher Models}

We compare four representative teacher models, KG-BERT (RoBERTa-large) and GPT2 (small, medium, and XL), on how pseudo labels provided by them can influence the model performance. The ablation results are shown in Table~\ref{table:ablation_teacher_model}. From the comparison between KG-BERT (RoBERTa-large, 340M parameters) and GPT2-small (117M) and -medium (345M), which are of the same scale of model size, we find that GPT2 can perform consistently better than KG-BERT as teacher models. This can be validated by the out-of-distribution performance in Table~\ref{table:overall_result} for the supervised learning baselines, where KG-BERT performs almost 3 points behind GPT2-medium in terms of OOD AUC. This ablation indicates the importance of powerful generalizable teacher models on pseudo labeling.

\finalcopy{
\subsubsection{Thresholding}\label{sec:threshold_ablation}
We study the sensitivity of thresholds in Figure~\ref{fig:threshold_ablation}. In this ablation, for simplicity, we set different $\mathcal{T}_{max}^+$ for positive pseudo examples, and sample the same amount of triples as the original training set whose $\alpha(x)<\mathcal{T}_{max}^+$ in descending order. We do the same ablation study on $\mathcal{T}_{min}^-$ for negative pseudo examples. 
When tuning one threshold, other thresholds are fixed as in section~\ref{sec:experimental_settings}.
The pseudo labels under different thresholds are directly used for PseudoReasoner without filtering, and we plot the test set AUC given different thresholds. We see that the resulting AUC is stable within certain ranges of $\mathcal{T}_{max}^+$ and $\mathcal{T}_{min}^-$. While when we set $\mathcal{T}_{min}^-$ to $-\infty$, which indicates no thresholds for negative examples are set, the performance drops drastically.
}

\subsubsection{Pseudo Label Filter}

\begin{figure}[t]
    \centering
    \includegraphics[width=0.8\linewidth]{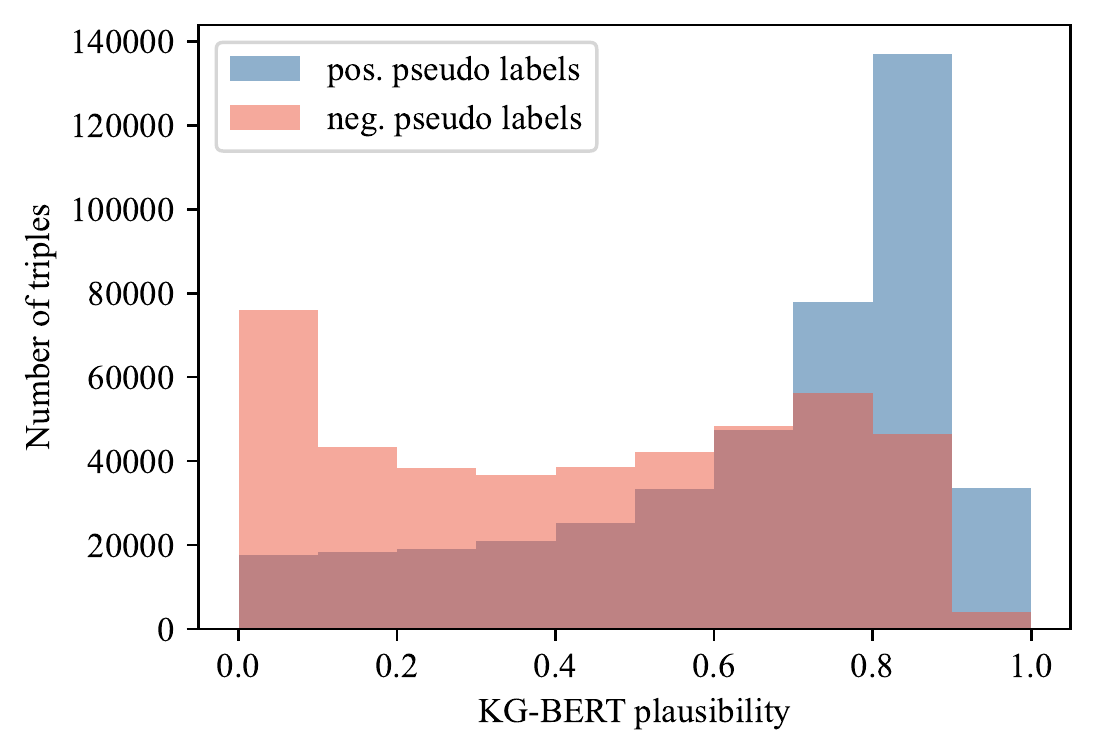}
    \vspace{-0.16in}
    \caption{KG-BERT plausibility distribution for positive/negative pseudo labels provided by GPT2.}
    \label{fig:kgbert_filter}
\end{figure}

We conduct experiments with different combinations of filtering mechanisms in Table~\ref{table:ablation_filter} for KG-BERT (RoBERTa-large). We can see that both filters (influence function and KG-BERT probability) benefits the model performance, while KG-BERT probability contributes to a more substantial improvement. 
Figure~\ref{fig:kgbert_filter} shows an illustration of the KG-BERT plausibility $\alpha(x)=P_{\theta}^*(y\text{=}1|x)$ of positive/negative pseudo examples provided by GPT2. The positive pseudo examples tend to be scored higher by KG-BERT than negative pseudo examples. Adding KG-BERT probability as an additional filter with the labels provided by GPT2 is similar to an ensembling procedure. 


\begin{table}[t]
\scriptsize
\centering
\begin{tabular}{c|cccc}
\toprule
& w/o filter & influence & KG-BERT & both\\
\midrule
UDA & 3.3 M & - & - & - \\ 
G-DAUG & 1.1M & 399.8K & 323.3K & 160.1K \\
PseudoReasoner & 932.6K & 408.5K & 373.0K  & 170.1K \\
Original & 1.1 M  & - & - & - \\
\bottomrule
\end{tabular}
\vspace{-0.1in}
\caption{ Number of pseudo examples used in experiments for semi-supervised methods. The ``Original'' row indicates the number of training examples in the original training set.}
\vspace{-0.2in}
\label{table:num_pseudo}
\end{table}

\subsection{Comparisons with Other Semi-supervised Learning Methods}\label{section:comparison}
\finalcopy{
\subsubsection{Computational Cost}
The number of pseudo examples used for semi-supervised-learning baselines are listed in Table~\ref{table:num_pseudo}. We basically use the same scale of unfiltered pseudo examples for G-DAUG and PseudoReasoner, while use 3 times more unlabeled examples for UDA as it requires more unlabeled data. 
Specifically, G-DAUG (COMET-distill) leverages the distilled knowledge from GPT3~\cite{DBLP:conf/naacl/WestBHHJBLWC22}, which further magnify the computational cost to more orders of magnitude.
In all, under the same scale of pseudo labels, PseudoReasoner can achieve far better results than UDA and G-DAUG.
}


\finalcopy{
\subsubsection{Analysis}
}
In UDA, though robustness can be improved with consistency loss on noised inputs, there are no \textit{new} commonsense knowledge added to the training procedure, making it hard for the model to be equipped with novel knowledge reasoning ability. 

For G-DAUG, the critial part lies in the generation of negative examples. We finetune two separate GPT2 on $D_l^+$ and $D_l^-$, and the one finetuned on $D_l^-$ is used to generate negative examples. 
Compared with the GPT2 finetuned on $D_l^+$, the GPT2 finetuned on $D_l^-$ is of a relatively lower quality as triples in $D_l^-$ don't follow specific commonsense patterns. 
We check the text generation quality of GPT2-XL finetuned on $D_l^+$ and $D_l^-$, and find that the BLEU-2 scores on two corresponding held-out test sets are 0.23 and 0.10, indicating the generation of negative examples are of a lower quality.

For Noisy-student, the main differences between our PseudoReasoner is that they use KG-BERT to provide pseudo labels, and they are soft pseudo labels. 
The main reason behind is that as soft labels are used, the teacher model has to be a discriminative model such as KG-BERT, which has a poor generalization ability than GPT2 used in our PseudoReasoner. 
Moreover, similar to the case in UDA, noisy is not a dominate factor in CSKB Population, while more high-quality novel commonsense knowledge matters more.

\begin{figure}[t]
    \centering
    \includegraphics[width=0.9\linewidth]{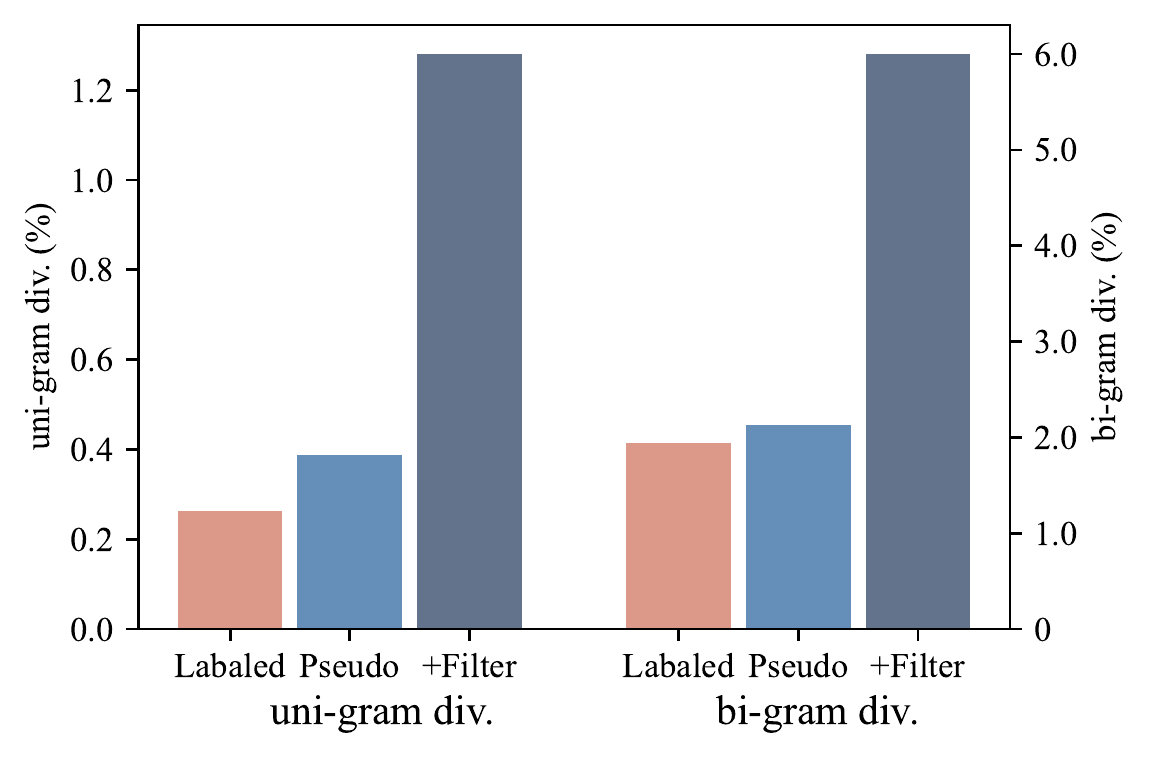}
    \vspace{-0.18in}
    \caption{Diversity analysis with the proportion of unique uni/bi-grams in the labeled dataset, the pseudo labels, and the filtered pseudo labels. With filtering, the diversity can be significantly improved.}
    \label{fig:diversity}
    \vspace{-0.2in}
\end{figure}



\subsection{Semantic Diversity Analysis } 

An important contribution of PseudoReasoner is that we extend the knowledge space for training from limited CSKBs to a more broad unlabeled resource. 
We use the proportion of unique uni-grams and bi-grams as an indicator of semantic diversity to measure the scale to which models are exposed to diverse novel knowledge. 
Figure~\ref{fig:diversity} shows that after filtering with influence function and KG-BERT probability, the diversity can be improved by around 3 to 4 times than the labeled dataset.


\subsection{Relationship with Knowledge Distillation} 
While knowledge distillation focuses on distilling knowledge from larger models to smaller ones, our method does not necessarily need the teacher model to be larger. The teacher GPT2-medium, which is of the same size as the student KG-BERT, can work pretty well and is comparable to GPT2-XL.

\section{Conclusion}

In this paper, we propose a semi-supervised learning framework for CSKB Population based on pseudo labels. Using a teacher model and a special filtering mechanism on pseudo labels, we achieve the state-of-the-art of CSKB Population in terms of both in-domain and out-of-domain performance. Experiments also show that our CSKB Population benefits more from high-quality novel knowledge than other semi-supervised learning techniques such as noise and consistency training. This work brings a new perspective of improving out-of-domain generalizable commonsense reasoning ability on CSKBs.

\section*{Acknowledgement}
\finalcopy{
The authors of this paper were supported by the NSFC Fund (U20B2053) from the NSFC of China, the RIF (R6020-19 and R6021-20) and the GRF (16211520) from RGC of Hong Kong, the MHKJFS (MHP/001/19) from ITC of Hong Kong and the National Key R\&D Program of China (2019YFE0198200) with special thanks to HKMAAC and CUSBLT, and the Jiangsu Province Science and Technology Collaboration Fund (BZ2021065). We thank the support from the UGC Research Matching Grants (RMGS20EG01-D, RMGS20CR11, RMGS20CR12, RMGS20EG19, RMGS20EG21). 
We also thank NVIDIA AI Technology Center (NVAITC) for the support of computational resources for conducting the experiments.
}

\section*{Limitations}

The main limitation is that for the labeling procedure of the teacher model, we still need a hard threshold $\mathcal{T}$ to determine whether the label is 0 or 1. 
This would limit the method from generalize across different tasks, where different thresholds should be tuned separately.
One solution and future work would be automatically learn the thresholding, or
improve the teacher-student interaction by training them jointly in an end-to-end manner such as Meta Pseudo Label~\cite{DBLP:conf/cvpr/PhamDXL21}. 

\bibliography{anthology,custom}
\bibliographystyle{acl_natbib}

\clearpage

\appendix

\section{Additional Details on the CSKB Population Benchmark} \label{appendix:ckbp_task}

Commonsense Knowledge Base (CSKB) Population aims at automatically populating commonsense knowledge defined in source CSKBs on an unlabeled candidate eventuality knowledge graph, ASER. We provide additional details about the selection of CSKBs and the eventuality knowledge graph, and the details about the evaluation of CSKB Population in this section.

\subsection{CSKB}

CSKB Population studies commonsense relations among general events. 
To form a general aligned CSKB, \citet{DBLP:conf/emnlp/Fang2021CKBP} selected ConceptNet~\cite{liu2004conceptnet} (the event-related relations are selected), ATOMIC~\cite{sap2019atomic}, ATOMIC$_{20}^{20}$~\cite{DBLP:conf/aaai/Hwang2021comet} (the newly added relations beyond ATOMIC are selected), and GLUCOSE~\cite{mostafazadeh2020glucose}.
These four sources of CSKBs are in the form of free-text and are structured as triplet forms ($(h, r, t)$).
Normalization processes are conducted to align the four CSKBs together.
The overall structure of the aligned CSKB is based on the formats in ATOMIC, where the events are person-centric sentences with \textit{PersonX} and \textit{PersonY} as subjects.
In ConceptNet, \textit{PersonX} is prepended on the \textit{predicate-object} pairs to make them complete sentences. For example, a triple in ConceptNet (lie, \texttt{HasSubEvent}, make up story) is converted to (PersonX lies, \texttt{HasSubEvent}, PersonX makes up story). 
The \textit{SomeoneA} and \textit{SomeoneB} are converted to \textit{PersonX} and \textit{PersonY} accordingly in GLUCOSE. The relationships in GLUCOSE are converted to ATOMIC formats according to the official conversion rules defined in Table 7 in \citet{mostafazadeh2020glucose}.

A summary of the relations studied in the aligned CSKBs is shown in Table~\ref{table:cskb_rel_stat}.

\subsection{ASER}

The candidate knowledge graph for populating commonsense knowledge,
ASER \cite{DBLP:conf/www/ZhangLPSL20aser, ZHANG2022103740aser}, is a large-scale eventuality-centric knowledge graph that provides explicit discourse relationships between eventualities.
The core part of ASER is used, where 10M discourse edges among 27M eventualities are included.
An example of an ASER edge is: (``I am hungry,'' \textit{Result}, ``I have lunch''), and such an discourse edge can be a potential \texttt{xWant} relation in CSKBs.
Such detailed conversion rules from discourse relations to commonsense relations are provided in \citet{DBLP:conf/emnlp/Fang2021CKBP}. Such conversion rules make the total number of unlabeled candidate edges around 200M.

\subsection{Evaluation Set}

For the ground truth commonsense triples from the CSKBs, they are split in to train, development, and test sets using the original split by their own papers. For example, in ATOMIC, the test set split is based on the skeleton words of the heads such that similar heads are split into the same dataset. The test set here is denoted as the \textit{original test set}.

The final evaluation set comprises of three parts, one sampled from the original automatically constructed test set as above (denoted as ``\textit{Original Test Set}''), one sample from the edges where  heads are from CSKBs and tails are from ASER (denoted as ``\textit{CSKB head + ASER tail}''), and one from ASER solely (denoted as ``\textit{ASER edges}'').
The total number of final evaluation set is around 32K, and they are manually annotated using Amazon Mechanical Turk.

\begin{table}[t]
\small
\centering
\begin{tabular}{p{1.5cm}|ccc}
\toprule
\scriptsize{Relation} & \scriptsize{ATOMIC($_{20}^{20}$)} & \scriptsize{ConceptNet} & \scriptsize{GLUCOSE} \\
\midrule
\scriptsize{\texttt{oEffect}} & 21,497 & 0 & 7,595 \\
\scriptsize{\texttt{xEffect}} & 61,021 & 0 & 30,596 \\
\scriptsize{\texttt{gEffect}} & 0 & 0 & 8,577 \\
\scriptsize{\texttt{oWant}} & 35,477 & 0 & 1,766 \\
\scriptsize{\texttt{xWant}} & 83,776 & 0 & 11,439 \\
\scriptsize{\texttt{gWant}} & 0 & 0 & 5,138 \\
\scriptsize{\texttt{oReact}} & 21,110 & 0 & 3,077 \\
\scriptsize{\texttt{xReact}} & 50,535 & 0 & 13,203 \\
\scriptsize{\texttt{gReact}} & 0 & 0 & 2,683 \\
\scriptsize{\texttt{xAttr}} & 89,337 & 0 & 7,664 \\
\scriptsize{\texttt{xNeed}} & 61,487 & 0 & 0 \\
\scriptsize{\texttt{xIntent}} & 29,034 & 0 & 8,292 \\
\scriptsize{\texttt{isBefore}} & 18,798 & 0 & 0 \\
\scriptsize{\texttt{isAfter}} & 18,600 & 0 & 0 \\
\scriptsize{\texttt{HinderedBy}} & 87,580 & 0 & 0 \\
\scriptsize{\texttt{xReason}} & 189 & 0 & 0 \\
\scriptsize{\texttt{Causes}} & 0 & 42 & 26,746 \\
\scriptsize{\texttt{HasSubEvent}} & 0 & 9,934 & 0 \\
\midrule
\scriptsize{Total}& 578,252 & 10,165 & 126,776 \\

\bottomrule
\end{tabular}
\caption{Relation distribution statistics for different CSKBs. The table is the same as in the Table 4 of \citet{DBLP:conf/emnlp/Fang2021CKBP}.
} \label{table:cskb_rel_stat}
\end{table}

\section{Influence Function} \label{appendix:influence_function}

Filtering out detrimental training examples with influence function~\cite{DBLP:conf/icml/KohL17} can boost the model performance, as shown in~\citet{DBLP:conf/emnlp/YangMFSBWBCD20GDAUG} and \citet{ DBLP:conf/acl/HanWT20}.
A training example $z=((h, r, t), y)$ will hurt the generalization ability of the model if including $z$ in the training set results in a higher validation loss. Denote $L(\mathcal{Z}, \theta)$ as the loss function of dataset $\mathcal{Z}$ under the parameter set $\theta$. Then the loss under the training set $\mathcal{Z}_{train}$ of cardinality $n$ is indicated in Equation~(\ref{eq:influence_loss}) in the main body (also shown below).

\begin{equation} \nonumber 
    L(\mathcal{Z}_{train}, \theta) = \frac{1}{n} \sum_{i=1}^{n} L(z_i, \theta).
\end{equation}

Denote $\theta^*$ as the optimized parameters after training the model on $\mathcal{Z}_{train}$, and $\theta^*_{-z}$ as the optimized parameters after training the model on $\mathcal{Z}_{train}-\{z\}$.
Denote $\mathcal{Z}_{val}$ as the validation set.
The empirical criterion to determine $z$ as a detrimental training example is Equation~(\ref{eq:influence_criterion}) in the main body (also shown below).
\begin{equation} \nonumber 
    L(\mathcal{Z}_{val}, \theta^*) - L(\mathcal{Z}_{val}, \theta^*_{-z})>0.
\end{equation}


Influence function gives an efficient approximation for the above quantity, whose idea is to compute the change of the optimized parameters thus the validation loss if $z$ were upweighted by a small $\epsilon$. It introduces new parameters $\theta^{*}_{\epsilon, z}= \arg \min_{\theta} \frac{1}{n} \sum_{i=1}^n L(z_i, \theta) + \epsilon L(z, \theta)$. \citet{zbMATH03899977} gives us the influence of upweighting $z$ on the optimized parameters $\theta^{*}$:
\begin{equation}\nonumber
    \mathcal{I}_{up, params}(z) = \frac{d\theta^{*}_{\epsilon, z}}{d\epsilon}\Big|_{\epsilon=0} = - H_{\theta^{*}}^{-1}\nabla_{\theta}L(z, \theta^{*})
\end{equation}
where $H_{\theta^{*}} = \frac{1}{n}\sum_{i=1}^{n} \nabla_{\theta}^2 L(z, \theta^{*})$ is the Hessian matrix. Thus, applying the chain rule, we calculate the influence of upweighting $z$ on the validation loss:
\begin{align*}
\mathcal{I}_{up, loss}(z) &= \frac{dL(\mathcal{Z}_{val}, \theta^*_{\epsilon,z})}{d\epsilon}\Big|_{\epsilon=0} \\
&= \nabla_{\theta} L(\mathcal{Z}_{val}, \theta^*)^\top \frac{d\theta^{*}_{\epsilon, z}}{d\epsilon}\Big|_{\epsilon=0} \\
&= - \nabla_{\theta} L(\mathcal{Z}_{val}, \theta^*)^\top H_{\theta^*}^{-1} \nabla_{\theta} L(z, \theta^*).
\end{align*}

Since removing $z$ is equivalent to unweighting it by $\epsilon = -\frac{1}{n}$, small when $n$ is large, we can linearly approximate $L(\mathcal{Z}_{val}, \theta^*) - L(\mathcal{Z}_{val}, \theta^*_{-z})$ in Equation (\ref{eq:influence_criterion}) as $-\epsilon\mathcal{I}_{up, loss}(z) \left(= \frac{1}{n}\mathcal{I}_{up, loss}(z)\right)$.

In practice, we also linearly approximate $\mathcal{I}_{up, loss}$ with inverse Hessian-vector product (HVP) introduced in LiSSA~\cite{DBLP:journals/jmlr/AgarwalBH17} following~\cite{DBLP:conf/icml/KohL17}.

\begin{figure}[t]
    \centering
    \includegraphics[width=1\linewidth]{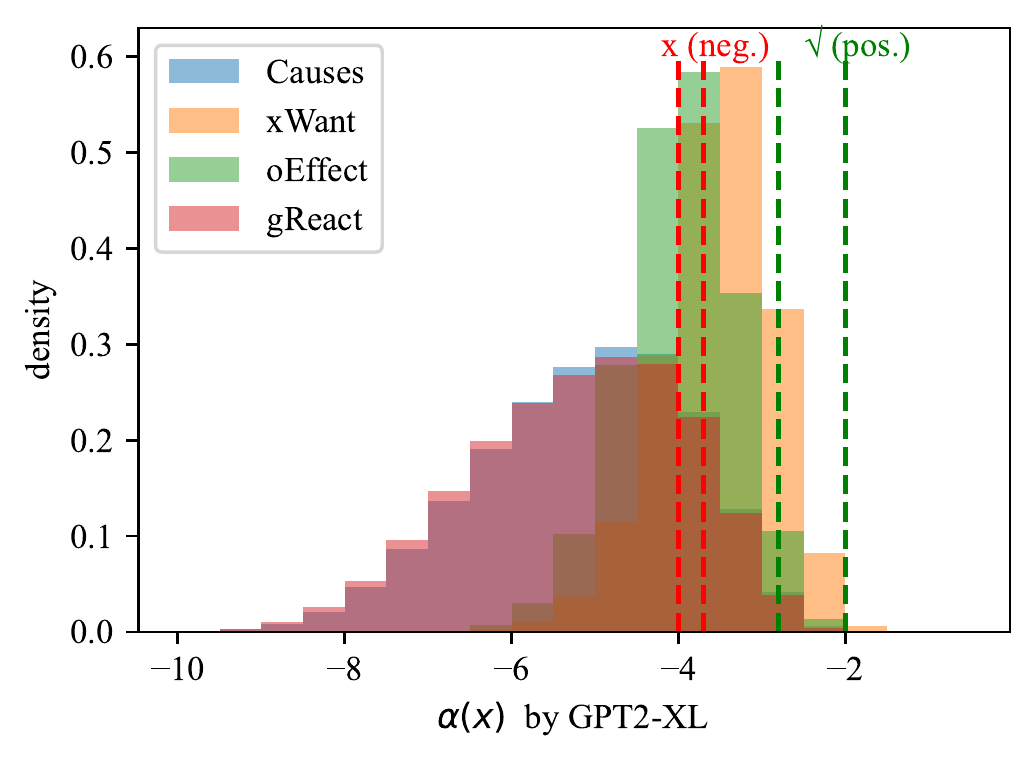}
    \vspace{-0.18in}
    \caption{GPT2-XL plausibility distribution ($-\alpha(x)$) of unlabeled data with relation \texttt{Causes}, \texttt{xWant}, \texttt{oEffect}, and \texttt{gReact}. The triples within the \Red{red} dashed lines are considered negative pseudo examples, and the triples within the \Green{green} dashed lines are considered as positive pseudo examples.}
    \label{fig:threshold_histogram}
\end{figure}

\section{Additional Details on Hyper-parameters}\label{appendix:hyper_params}

We present additional details about hyperparameters in the thresholding of the GPT2-XL  teacher model and the details of the backbone models.

\subsection{Thresholding in Pseudo Labels}

We present the GPT2-XL plausibility scores on the unlabaled $D_u$ in Figure~\ref{fig:threshold_histogram} for four representative commonsense relations, \texttt{Causes}, \texttt{xWant}, \texttt{oEffect}, and \texttt{gReact}. 
We set the thresholds $\mathcal{T}^-_{min}\text{=}-4.0, \mathcal{T}^-_{max}\text{=}-3.7, \mathcal{T}^+_{min}\text{=}-2.8, \mathcal{T}^+_{max}\text{=}-2.0$, by roughly observing the data distribution and representative triples in different range of plausibility scores.
\finalcopy{The filtered triples are further sampled such that the number of pseudo examples per relation equals the number of the original examples for the corresponding relations.}
\finalcopy{The pseudo examples that we use for experiments can be downloaded at our GitHub repository.}

\subsection{Model Training Details}\label{appendix:training_detail}

For KG-BERT, when BERT, RoBERTa, and DeBERTa is used,  an $(h, r, t)$ triple is converted to ``$\text{[CLS]}, h_1, ..., h_{|h|}, \text{[SEP]}, \text{[$r$]}, \text{[SEP]}, t_1, ..., t_{|t|}$''. Here, [CLS] and [SEP] are the special tokens in BERT-based models that are used to represent the whole sentence, and the special token to separate different sentences, respectively. $h_1, ..., h_{|h|}$ is the tokenized tokens of the head $h$, and  $t_1, ..., t_{|t|}$ is the tokenized tokens of the tail $t$. [$r$] is registered as a new special token for a certain relation $r$. 
The embedding of [CLS] in the last hidden layer of the BERT-based encoder is used as the final representation of the whole triple. 

When BART is used as the the backbone, an $(h, r, t)$ triple is converted to ``$\text{<s>}, h_1, \cdots, h_{|h|}, \text{[}r\text{]}, t_1, \cdots, t_{|t|}, \text{</s>}$'', where <s> and </s> are the special tokens in BART to represent starting-of-sentence and end-of-sentence. The sentence is fed into the encoder of BART and also the decoder of BART. The embedding of the end-of-sentence token </s> in the last hidden layer of the decoder is used as the embedding of the whole triple. This method is the default usage as in the original paper of BART for text classification.

For GPT2, we adopt the text generation setting in COMET~\cite{DBLP:conf/acl/Bosselut2019comet}, where the task is defined as give the head and relation to generate tail. Specifically, an $(h, r, t)$ triple is converted to ``$\text{[SOS]}, h_1, \cdots, h_{|h|}, \text{[}r\text{]}, t_1, \cdots, t_{|t|}, \text{[EOS]}$'', where [SOS] and [EOS] are special tokens representing start-of-sentence and end-of-sentence.
The negative GPT2 loss of the whole sentence (Equation~(\ref{eq:gpt2loss})) is used as the plausibility of the triple. 
Such a GPT2 can also be used to generate positive commonsense tails in the G-DAUG baseline.
We adapt the codes$\footnote{https://github.com/allenai/comet-atomic-2020}$ from ATOMIC$_{20}^{20}$ to our code base.

We build all models based on PyTorch and all transformer models from Hugginface transformers.
For training, we use 1e-5 as the default learning rate for all training stages. The batch size is set to 64 for all experiments. 
The maximum sequence length is set to 30, which can cover over 99\% of the serialized triples.
We use ADAM as the optimizer to train all models.
We evaluate the validation AUC every 250 steps and store the checkpoint when the largest validation AUC is achieved. 
We run all experiments with three random seeds (100, 101, and 102) and report the average.
The KG-BERT related experiments are run on NVIDIA RTX 3080 Ti, and the GPT2 related experiments are run on NVIDIA A40.

\section{Details of Baselines} \label{sec:appendix:baseline}

We present more details for the training of semi-supervised learning baselines in this section.

\subsection{UDA}\label{appendix:uda}

In the original paper of UDA, two data augmentation methods, back-translation and TF-IDF replacement, are introduced for natural language tasks. We use both augmentation methods and compare the results. 
For each augmentation methods, original serialized unlabeled candidates $x$ are duplicated and augmented, then both original and augmented candidates are used for consistency training. 

For back-translation, we employ OPUS translation models\footnote{https://huggingface.co/Helsinki-NLP} between English and French. Specifically, in each candidate, the ``PersonX'' is replaced with Alice, ``PersonY'' with Charles, and ``PersonZ'' with Francisco, then candidates are forward-translated to French and after then backward-translated to English. Selected names Alice, Charles, and Francisco are common first names shared by French and English and remain unchanged after back-translation, thus it ensures that the personal pronouns PersonX, PersonY, and PersonZ in unlabeled candidates can be fully recovered after back-translation. Also, we observe that almost newly registered relation tokens survive after the back-translation process, bringing the opportunity for KG-BERT to get familiar to these tokens in consistency training. 
For TF-IDF replacement, we set the replacement probability as 0.1 to avoid cracking the main semantics of the original sentences. 


We present more hyperparameter tuning here for UDA. The objective function of UDA is as follows in Equation~(\ref{eq:uda}), where $p_L$ and $p_U$ are the distribution of the labeled and unlabeled dataset, $y_1$ is the corresponding label of $x_1$, $p_\theta$ is the predicted probability by the backbone model under parameter $\theta$, $q(\hat{x}|x_2)$ is a data augmentation transformation, $\Tilde{\theta}$ is a fixed copy of the current parameters $\theta$, and CE indicates cross entropy. $\lambda$ is used to control the weight of unsupervised consistency loss. Besides $\lambda$, in the training process, there is another parameter $r$ indicating the ratio of unsupervised examples in a batch. 
We conduct several experiments with grid search to find the best combinations of $\lambda$ and $r$ as shown in Table~\ref{table:uda_grid} when using TF-IDF replacement as the augmentation method. 
We report the results of the best combinations in the main text of the paper.


\begin{equation}\label{eq:uda}
    \begin{aligned}
        &J(\theta) = \mathbb{E}_{x1\sim p_L(x)} [-\log p_\theta(y_1|x_1)]\\
        &+\lambda \mathbb{E}_{x_2\sim p_U(x)}\mathbb{E}_{\hat{x}\sim q(\hat{x}|x_2)}[\text{CE}(p_{\Tilde{\theta}}(y|x_2)||p_{\theta}(y|\hat{x})]
    \end{aligned}
\end{equation}



\begin{table}[t]
\small
\begin{center}
\begin{tabular}{c|c|c|c}
    \toprule
    \diagbox{\small{$r$}}{\small{$\lambda$}}&  0.3 & 1 & 3 \\ 
    \midrule
    0.3 & 71.0 & 71.4 & 71.2 \\
    1   & 71.4 & \textbf{71.7} & 71.2 \\
    3   & 71.2 & 71.2 & 71.4 \\
    \bottomrule
    
\end{tabular}
\caption{
Overall AUC score with respect to different pairs of unsupervised example ratio $r$ and consistency loss coefficient $\lambda$.
}
\label{table:uda_grid}
\end{center}
\end{table}



The original implementation of UDA is based on Tensorflow\footnote{https://github.com/google-research/uda}. We re-implement their code with pytorch and aggregate it into our code base. 

\subsection{Noisy Student}\label{appendix:noisy_student}

For Noisy Student, they use a teacher model finetuned on the labeled dataset first to provide initial pseudo labels. 
Then student models are finetuned with both the labeled dataset the pseudo labels iteratively when the latest pseudo labels are provided by the latest student model. 
Specifically, noise is added to the input during the stage for student models' training.

In this section, we provide some detailed ablation studies on the setting of Noisy Student.
First, we study the role of noise in Noisy Student in Table~\ref{table:noisy_student_detail} when noise is added or removed to the output logits of the model at the first training iteration. 
Results show that there can be a 0.3 points improvement on the overall AUC when dropout is added with probability either 0.1 or 0.5, showing that noise can indeed force the models to learn harder. 
Second, we study the performance across different iterations of Noisy Student. In table~\ref{table:noisy_student_detail}, the performance reaches the peak at iteration 3, indicating a balance of the student model's learning ability and generalizable pseudo labeling ability. We report the best result in the result table of the main body.

\subsection{G-DAUG}\label{appendix:gdaug}

For Generative Data Augmentation, we check the performance under different sizes of COMET (use GPT2-small and GPT2-XL as representatives) in Table~\ref{table:gdaug_detail}. 
Also, we check the three filtering mechanisms, influence function, diversity, and KG-BERT probability. The diversity filter is a heuristic algorithm that includes diverse training examples one by one as defined in the original paper.
We find that still the KG-BERT filter is the most effective filter and can boost the performance more than the other two. 
Also, G-DAUG also benefits more from a more powerful text generation model (GPT2-XL).

\begin{table}[t]
\small
\centering
\begin{tabular}{l|c|ccc}
\toprule
  \multicolumn{1}{l|}{}& \tabincell{c}{all} & \tabincell{c}{\scriptsize{\textit{Original}}\\ \scriptsize{\textit{Test Set}}} & \tabincell{c}{\scriptsize{\textit{CSKB head} }\\ \scriptsize{\textit{+ ASER tail}}} & \tabincell{c}{\scriptsize{\textit{ASER}}\\ \scriptsize{\textit{edges}}} \\
\midrule
\multicolumn{1}{l|}{\tabincell{l}{KG-BERT\\(RoBERTa-large)}}  & 70.9 & 78.0 & 63.4 & 64.6\\
\midrule
1st Iteration \\
\hline
\ - w/o noisy & 71.8 & 78.5 & 64.4 & 65.8 \\
\ - dropout ($p$=0.1) & 72.1 & 78.8 & 64.8 & 66.6 \\ 
\ - dropout ($p$=0.5) & 72.1 & 79.3 & 64.7 & 66.2 \\ 
\midrule
\ Num of Iterations \\
\hline
\qquad \qquad 1 & 72.1 & 79.3 & 64.7 & 66.2 \\
\qquad \qquad 2 & 72.1 & 78.6 & 64.6 & \textbf{66.8} \\
\qquad \qquad 3 & \textbf{72.4} & 79.3 & \textbf{65.3} & 66.7 \\
\qquad \qquad 4 & 72.3 & \textbf{79.4} & 65.1 & 66.4 \\
\bottomrule
\end{tabular}
\caption{ Ablation study on the effect of noisy in the Noisy Student model. $p$ means the dropout probability. }
\label{table:noisy_student_detail}
\end{table}

\begin{table}[t]
\small
\centering
\begin{tabular}{lc|ccc}
\toprule
  \multicolumn{1}{l|}{}& \tabincell{c}{all} & \tabincell{c}{\scriptsize{\textit{Original}}\\ \scriptsize{\textit{Test Set}}} & \tabincell{c}{\scriptsize{\textit{CSKB head} }\\ \scriptsize{\textit{+ ASER tail}}} & \tabincell{c}{\scriptsize{\textit{ASER}}\\ \scriptsize{\textit{edges}}} \\
\midrule
\multicolumn{1}{l|}{\tabincell{l}{KG-BERT\\(RoBERTa-large)}}  & 70.9 & 78.0 & 63.4 & 64.6\\
\midrule
\multicolumn{2}{l}{\textbf{G-DAUG (GPT2-small)}} \\
\hline
 \multicolumn{1}{l|}{\ no filter} & 71.3 & 78.6 & 63.7 & 65.3 \\
\multicolumn{1}{l|}{\ + influence} & 70.6 & 76.8 & 63.0 & \textbf{66.5} \\
\multicolumn{1}{l|}{\ + diversity} & 71.4 & 78.5 & 64.2 & 65.3 \\
\multicolumn{1}{l|}{\ + KG-BERT} &  \textbf{71.5} &\textbf{78.7} & \textbf{64.4} & 64.9 \\
\midrule
\multicolumn{2}{l}{\textbf{G-DAUG (GPT2-XL)}} \\
\hline
\multicolumn{1}{l|}{\ no filter} & 71.5 & 78.7 & 64.2 & 65.4\\
\multicolumn{1}{l|}{\ + influence} & 71.3 & 78.4 & 64.4 & 65.2 \\
\multicolumn{1}{l|}{\ + diversity} & 71.7 & \textbf{78.7} & 64.7 & 65.5 \\
\multicolumn{1}{l|}{\ + KG-BERT} & \textbf{71.7} & 78.5 & \textbf{64.8} & \textbf{65.5}\\
\bottomrule
\end{tabular}
\caption{ Effects of different filtering methods of the G-DAUG~\cite{DBLP:conf/emnlp/YangMFSBWBCD20GDAUG} baseline. We use RoBERTa-large as the base language model for KG-BERT. For example, ``+ KG-BERT'' means filter the augmented dataset with KG-BERT loss as illustrated in Section~\ref{sec:filter}. }
\label{table:gdaug_detail}
\end{table}




\end{document}